\documentclass[conference]{IEEEtran}
\usepackage{graphicx}
\usepackage{url}

\begin{document}

\title{The UMCD Dataset}

\author{
	\IEEEauthorblockN{Danilo Avola\IEEEauthorrefmark{1}, Gian Luca Foresti\IEEEauthorrefmark{1}, Niki Martinel\IEEEauthorrefmark{1}, Daniele Pannone\IEEEauthorrefmark{2} and Claudio Piciarelli\IEEEauthorrefmark{1}}
	\IEEEauthorblockA{\IEEEauthorrefmark{1}Department of Mathematics, Computer Science and Physics, University of Udine, Via delle Scienze 206, 33100 Udine, Italy
		}
	\IEEEauthorblockA{\IEEEauthorrefmark{2}Department of Computer Science, Sapienza University, Via Salaria 113, 00198 Rome, Italy
	}
}

\maketitle

\begin{abstract}
In recent years, the technological improvements of low-cost small-scale Unmanned Aerial Vehicles (UAVs) are promoting an ever-increasing use of them in different tasks. In particular, the use of small-scale UAVs is useful in all these low-altitude tasks in which common UAVs cannot be adopted, such as recurrent comprehensive view of wide environments, frequent monitoring of military areas, real-time classification of static and moving entities (e.g., people, cars, etc.). These tasks can be supported by mosaicking and change detection algorithms achieved at low-altitude. Currently, public datasets for testing these algorithms are not available. This paper presents the UMCD dataset, the first collection of geo-referenced video sequences acquired at low-altitude for mosaicking and change detection purposes. Five reference scenarios are also reported.
\end{abstract}
\section{Introduction}
Recently, the use of medium and large UAVs in civilian and military activities is increased considerably~\cite{rs4061671,6705275,6722317}. Usually, these UAVs acquire video sequences at very high-altitude (e.g., above 100 meters) for large-scale operations, such as vegetation monitoring and mapping of buildings. In a wide range of applications, including surveillance~\cite{Meng,Price}, search and rescue~\cite{rs9020100,s16111778,ROB:ROB21615,Piciarelli2013}, tracking~\cite{Zhang2016,s16091406,Breckon}, and many others, often video sequences acquired at very low-altitude are needed. In particular, these sequences are required in all those application contexts in which a frequent or continuous checking of an area of interest is necessary. Mosaicking and change detection techniques~\cite{Wischounig-Strucl2015,7817860,Avola2016} can support this type of tasks to detect, track, and classify static and moving entities on the ground. Currently, public datasets for testing mosaicking algorithms contain sequences acquired at very high-altitude, while, datasets that contain geo-referenced sequences acquired at low-altitude for testing change detection algorithms are not available. This paper proposes the UAV Mosaicking and Change Detection (UMCD) dataset to fill this lack. The dataset is composed of two main sets of challenging video sequences acquired at very low-altitude. The first set consists of 30 not geo-referenced sequences that can be used to evaluate mosaicking algorithms. The second set is made up of 10 pairs of geo-referenced sequences (i.e., 20 videos) in which the first can be used to build the mosaic and the second, acquired on the same path, can be used to test change detection algorithms. The geo-referencing allows developers to reduce drastically the number of matching during the search of entities. The UMCD dataset is freely available only for research purposes at the following link:~\url{www.umcd-dataset.net}.

The paper is structured as follows. Section~\ref{second} reports some examples of possible scenarios in which the dataset can be used. Finally, Section~\ref{third} concludes the paper.
\section{Potential Scenarios}
\label{second}
The UMCD dataset is designed to support different tasks for mosaicking and change detection purposes. Anyway, due to both the suitable length of the video sequences and the different entities contained in them, the sequences can be also adopted for different aims, e.g., object classification without using a reference mosaic. In this section, five scenarios that show the potential use of the dataset are presented. The dataset is built to allow the development of real-time and on-line algorithms, but this last aspect depends on the specific algorithm implementation. In the next sub-sections, the following five scenarios are shown: object detection, people search and rescue, people and vehicle classification, military camp monitoring, and urban area monitoring. 
\subsection{Scenario 1: Object Detection}
\begin{figure}[h]
	\centering
	\includegraphics[width=\columnwidth]{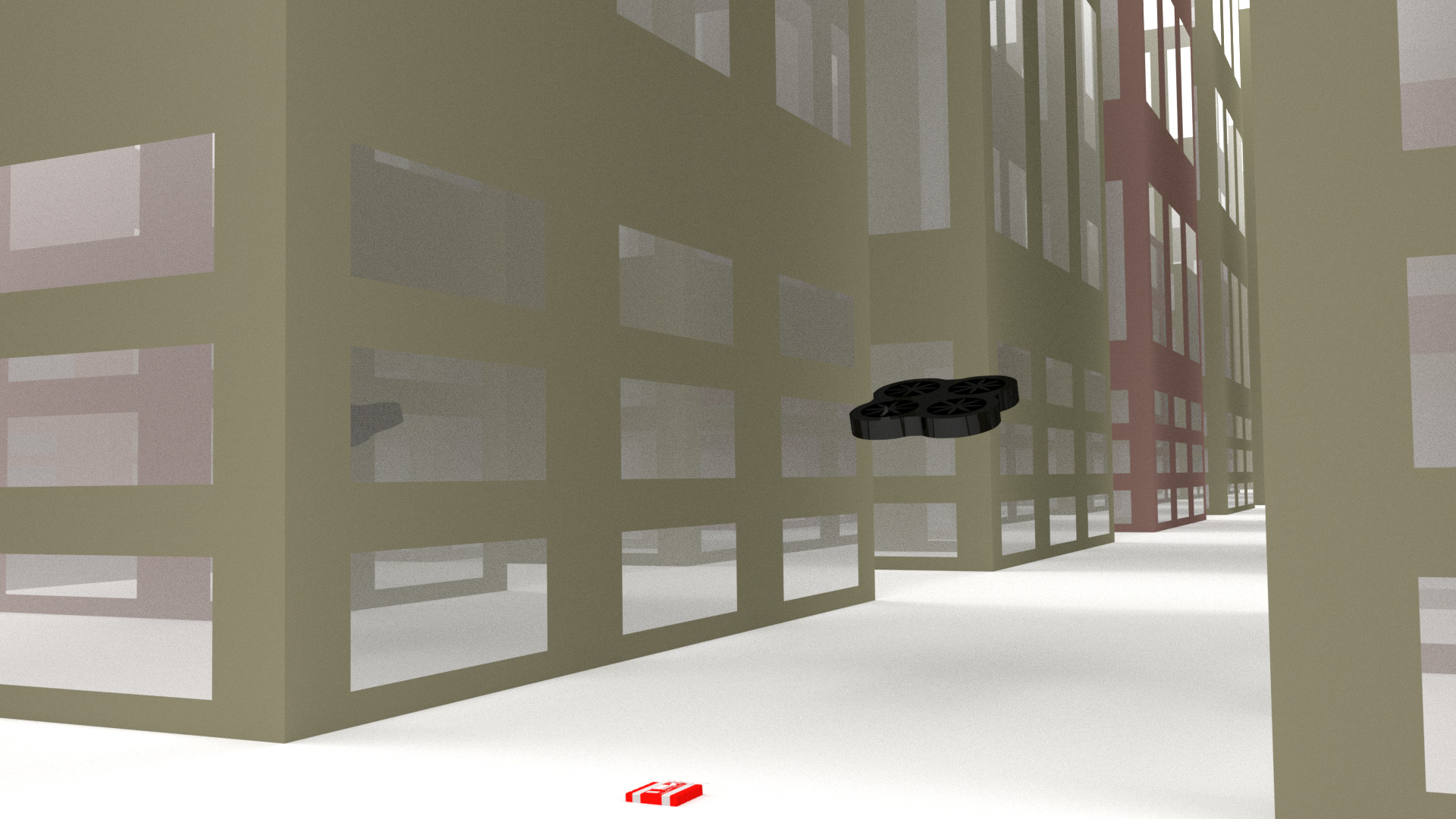}
	\caption{Scenario 1: Object detection.}
	\label{fig:case1}
\end{figure}

In the first scenario, the object detection task is reported (Figure~\ref{fig:case1}). Small objects, such as packets, boxes, and bags can be used to host dangerous ordnances, including Improvised Explosive Devices (IED). Due to the size of these objects, a high spatial resolution is required. Some medium and large UAVs can be equipped with very high resolution cameras, however due to their telemetry (e.g., speed, high-altitude) these aerial vehicles could not be suitable to detect small objects. Moreover, it might be necessary to perform such an object detection task frequently over time (e.g., several time in an hour), thus making impossible the use of common UAVs.
\subsection{Scenario 2: People Search and Rescue}
\begin{figure}[h]
	\centering
	\includegraphics[width=\columnwidth]{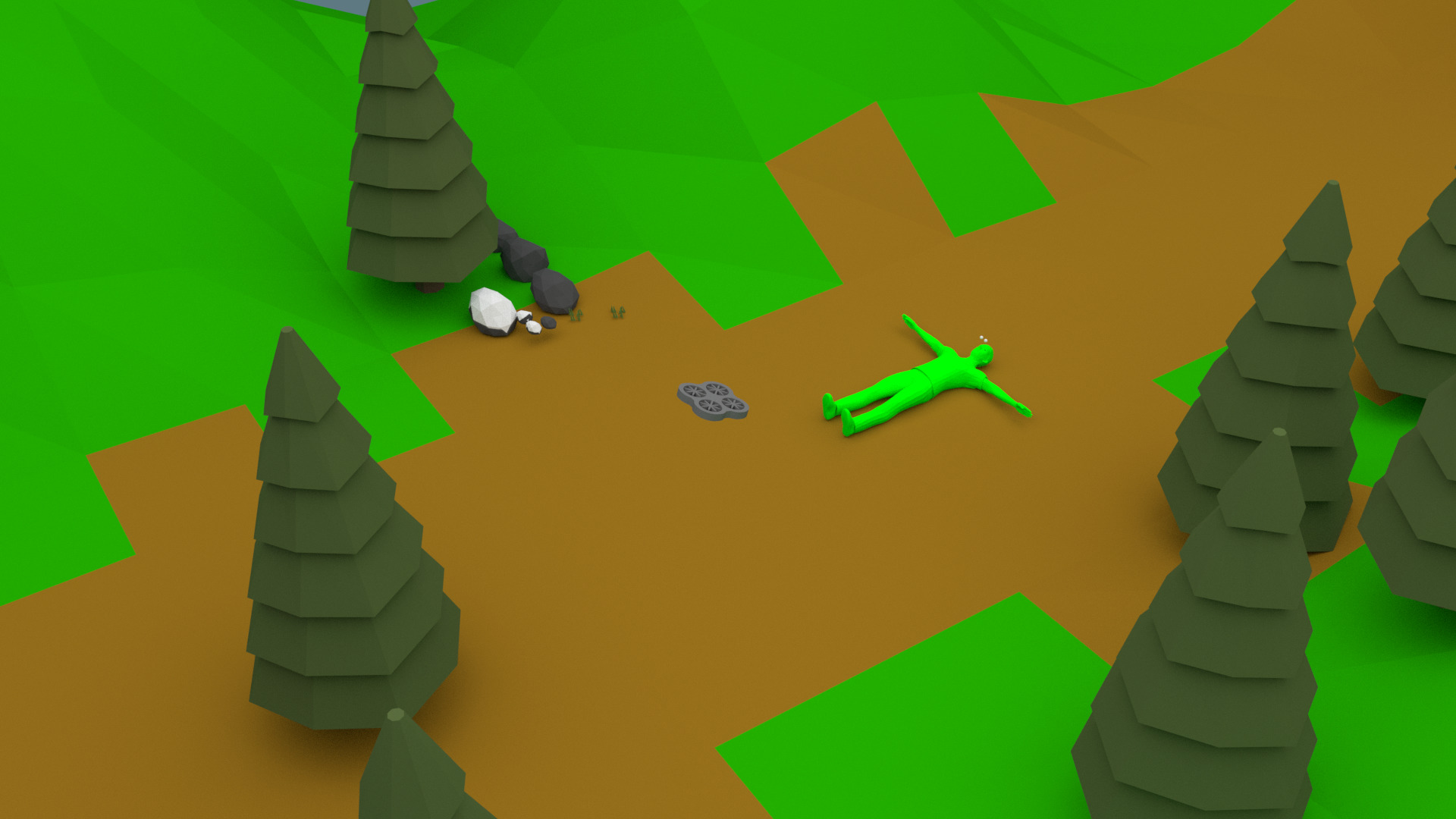}
	\caption{Scenario 2: People Search and Rescue.}
	\label{fig:case2}
\end{figure}

In the second scenario, the people search and rescue task is reported (Figure~\ref{fig:case2}). Often, this type of tasks occur in unfriendly areas (e.g., post-earthquakes, post-avalanche). The small-scale UAVs are agile and fast devices, moreover they can be designed to have autonomous behaviours (e.g., avoid obstacles, research a specific target), thus making them the most suitable aerial vehicles to successfully pursue such duties.
\subsection{Scenario 3: People and Vehicle Classification}
\begin{figure}[h]
	\centering
	\includegraphics[width=\columnwidth]{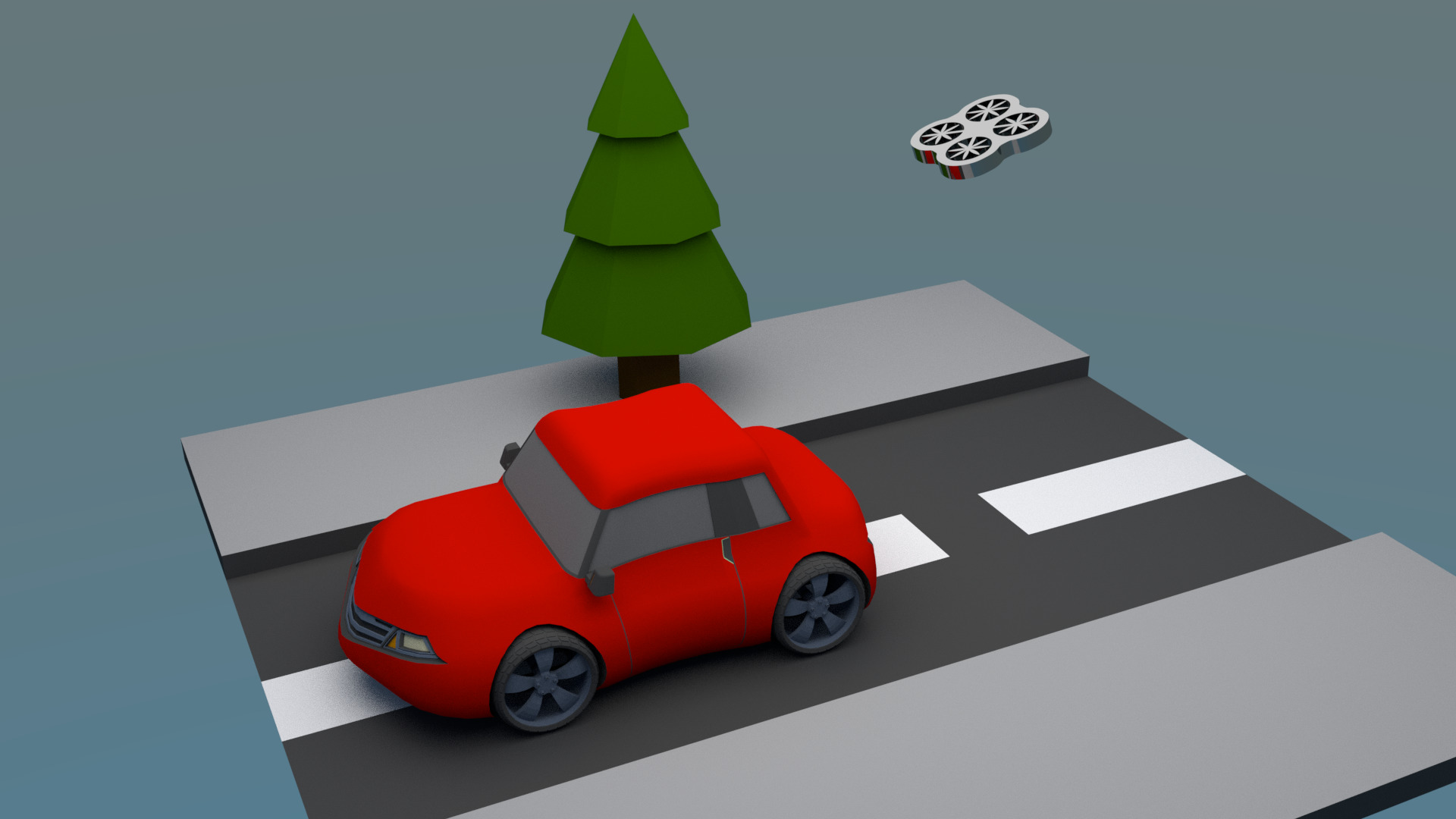}
	\caption{Scenario 3: People and Vehicle Classification.}
	\label{fig:case3}
\end{figure}

In the third scenario, the people and vehicle classification task is reported (Figure~\ref{fig:case3}). In this type of tasks the spatial resolution plays a main role. In fact, as well known, the spatial resolution required to classify an object must be higher with respect to that required to detect it. For this reason, the use of small-scale UAVs whose telemetry can be adapted according to the specific situations must be considered an optimal solution. The classification tasks cover a wide range of practical applications in safety and security, some examples include the access monitoring to quarantine zones, the surveillance of critical areas (e.g., border areas), the detection of intruders in off-limit zones, and many others. Moreover, these tasks can be considered a mandatory prerequisite for several complex duties, such as target tracking, re-identification, and others.
\subsection{Scenario 4: Military Camp Monitoring}
\begin{figure}[h]
	\centering
	\includegraphics[width=\columnwidth]{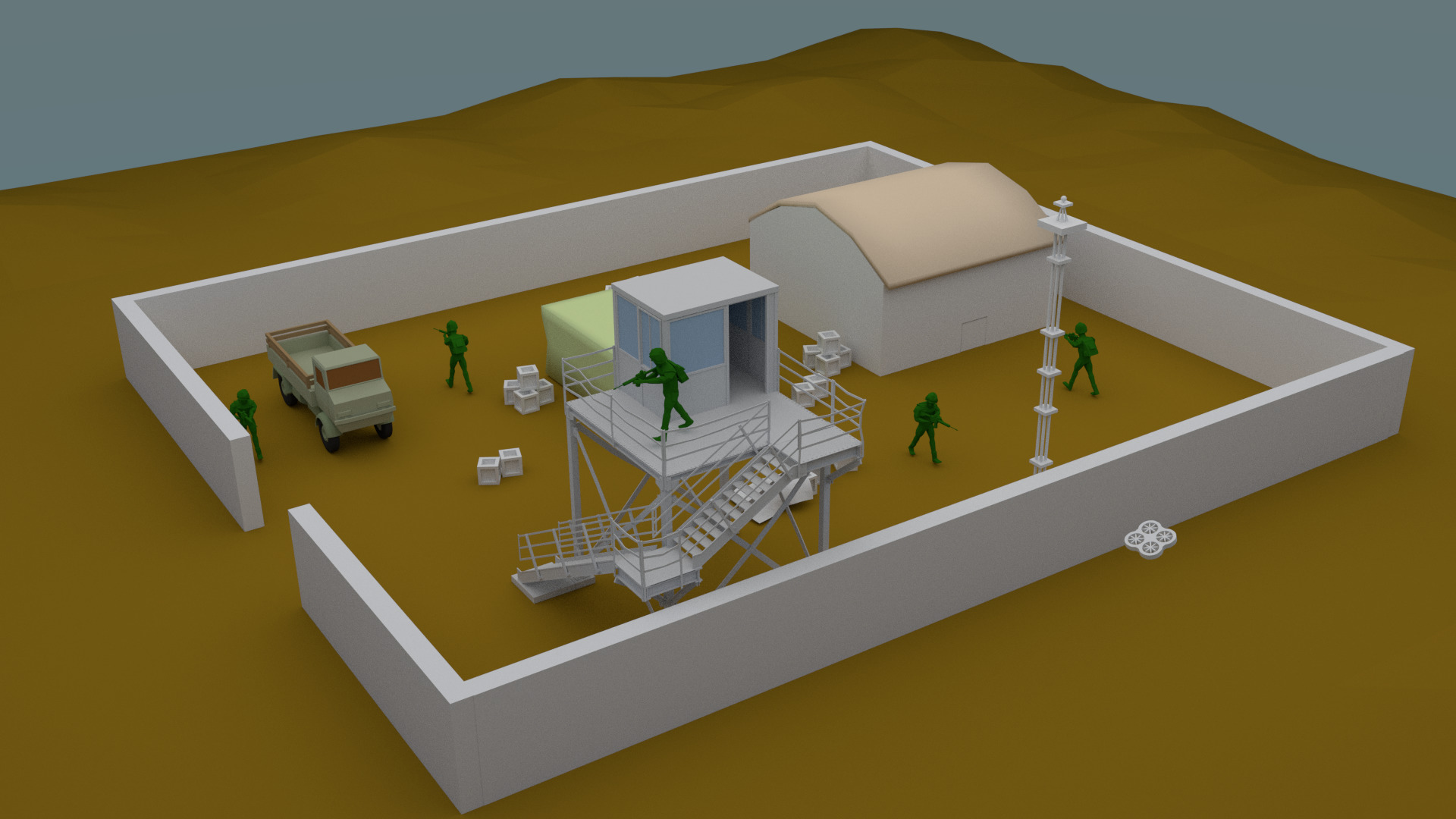}
	\caption{Scenario 4: Military Camp Monitoring.}
	\label{fig:case4}
\end{figure}

In the fourth scenario, the military camp monitoring task is reported (Figure~\ref{fig:case4}). In this task, the main goal is to check if outside the walls (or protections) some objects are suddenly appeared. Due to both the variability of the possible small objects and the fast changes of the surrounding environment, in this case the mosaicking and change detection approach is without doubt the best solution. In this specific case, a continuous monitoring could be necessary and a turnover of more small-scale UAVs could be adopted.
\subsection{Scenario 5: Urban Area Monitoring}
In the fifth, and last, scenario, the urban area monitoring task is reported. In several cases, restricted urban areas can be established for different reasons, including population control, looting monitoring, intrusion detection, and many others. These are other examples in which the feedback from a small-scale UAV to a human operator must be in real-time, thus promoting, once again, the use of these devices instead of common UAVs.
\begin{figure}[h]
	\centering
	\includegraphics[width=\columnwidth]{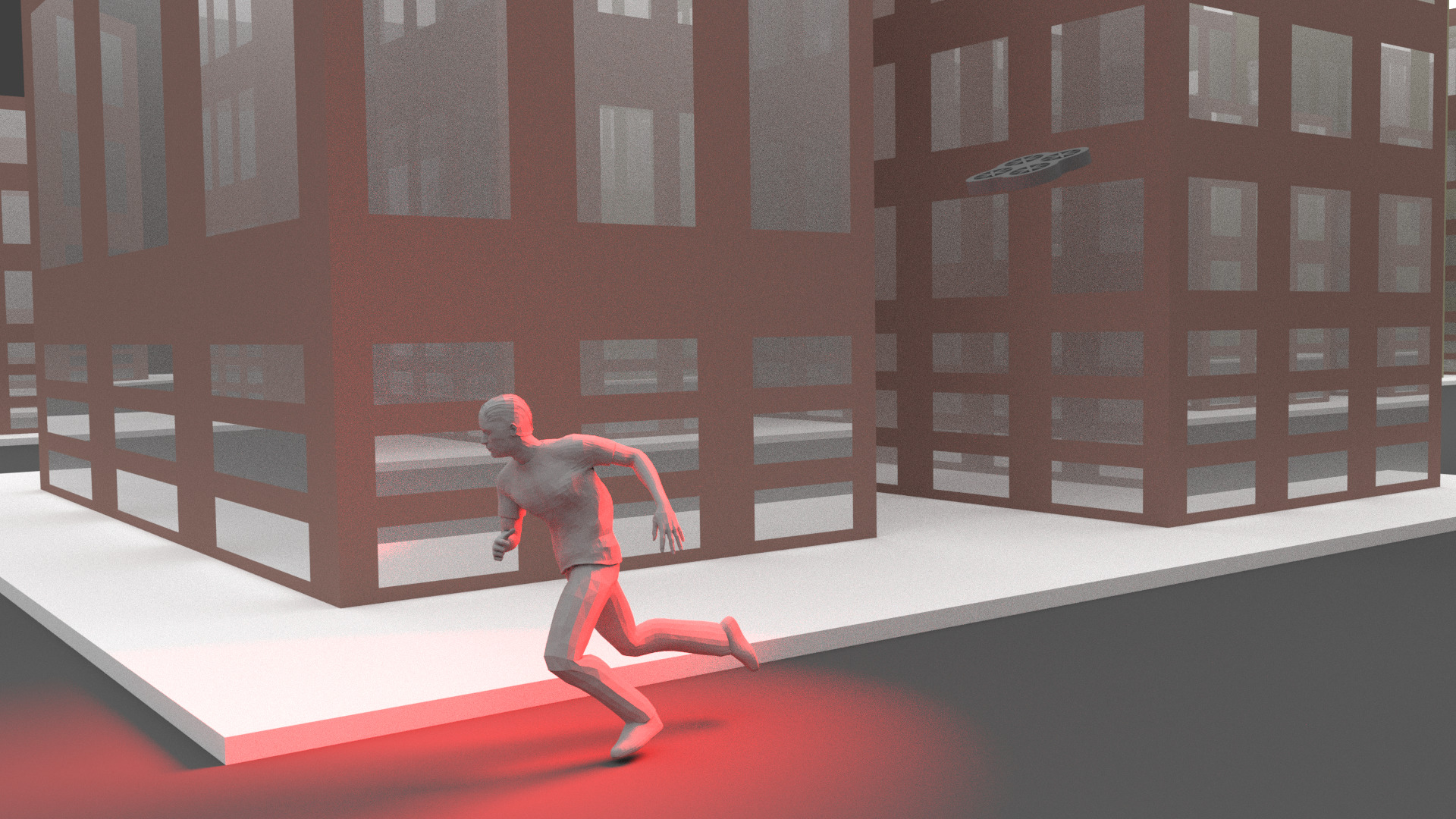}
	\caption{Scenario 5: Urban Area Monitoring.}
	\label{fig:case5}
\end{figure}
\section{Conclusions}
\label{third}
In the last years, the use of small-scale UAVs is increased greatly. Unlike the common medium or large aerial vehicles, the small-scale UAVs are extremely versatile, quickly programmable and very agile. In particular, they are highly suitable to perform tasks at very low-altitude. Currently, public aerial video sequences to test mosaicking algorithms are acquired at very high-altitude, while public geo-referenced sequences acquired at low-altitude to test mosaicking and change detection algorithms are not available. In this paper, the UMCD dataset is presented. The purpose of the dataset is to fill the previously highlighted lack. The paper also introduces five scenarios that suggest possible use of the dataset.
%
\ifCLASSOPTIONcaptionsoff
  \newpage
\fi

\bibliographystyle{IEEEtran}

\bibliography{references}

\end{document}